OPTIMIZATION

# Unknown area exploration for robots with energy constraints using a modified Butterfly Optimization Algorithm

Amine Bendahmane[1] · Redouane Tlemsani[1]

**Abstract**

Butterfly Optimization Algorithm (BOA) is a recent metaheuristic that has been used in several optimization problems. In this paper, we propose a new version of the algorithm (xBOA) based on the crossover operator and compare its results to the original BOA and 3 other variants recently introduced in the literature. We also proposed a framework for solving the unknown area exploration problem with energy constraints using metaheuristics in both single- and multi-robot scenarios. This framework allowed us to benchmark the performances of different metaheuristics for the robotics exploration problem. We conducted several experiments to validate this framework and used it to compare the effectiveness of xBOA with well-known metaheuristics used in the literature through 5 evaluation criteria. Although BOA and xBOA are not optimal in all these criteria, we found that BOA can be a good alternative to many metaheuristics in terms of the exploration time, while xBOA is more robust to local optima; has better fitness convergence; and achieves better exploration rates than the original BOA and its other variants.

## 1 Introduction

Unknown area exploration is one of the most active research topics in robotics. The goal is to deploy an autonomous robot inside an area to discover useful information or to locate a certain object/person.

Using robots for this type of task creates the opportunity to remotely study a dangerous zone without harming human operators or to delegate tedious missions to the machines such as creating the map of a big area, cleaning a house or a factory, locating gas leaks, landmines, or potential intruders inside a wide area. These kinds of tasks require decision-making efforts when performed by humans, which might produce different decisions according to their strategies or their level of expertise. Consequently, it is difficult to reproduce the process using traditional programming techniques based on conditional statements or using inference-based systems.

Robotics area exploration tasks can be divided in two types: known area exploration and unknown area exploration. In the latter variant, the robot has no prior information about the area to explore, which adds a level of difficulty to the task since the robot must adapt its decisions to all situations.

The goal of robotics exploration is to maximize the surface of the explored area while minimizing energy consumption. It is necessary to add some constraints in certain scenarios like keeping a communication range with a central station, avoiding unsafe situations (fire, water, etc.), or keeping a minimal distance from obstacles and humans. The whole process is then to find the right tradeoff between these criteria, which motivates the representation of this task as an optimization problem.

Since the search space grows exponentially relative to the size of the map to explore, it is not always possible to use a deterministic algorithm for finding the exact solution to this problem. A more practical solution is to use metaheuristics which proved to be efficient in finding acceptable solutions for large search spaces during a small amount of time.

✉ Amine Bendahmane
amine.bendahmane@univ-usto.dz

Redouane Tlemsani
redouane.tlemsani@univ-usto.dz

1 Signal-Image-Parole (SIMPA) Laboratory, Computer Science Department, Université des Sciences et de la Technologie d'Oran Mohamed Boudiaf, El Mnaouar, BP 1505, Bir El Djir, 31000 Oran, Algeria

A popular approach in the literature is to model robotics exploration problems as a traveling salesman problem where the robot must visit all regions of the area then return to the starting position. The solution will consist then of calculating the Hamiltonian path, which is the optimal path to visit every region once and only once. However, this requires knowing in advance the environment's structure and all the possible routes, which is not the case in the unknown area exploration missions.

Therefore, the robot has to start moving in a random direction then keeps updating its path continuously as new routes are discovered. It may happen that these routes change because of moving obstacles in the robot's environment (e.g.: chairs, doors, etc.). This means that the search space is subject to modifications which may change the configuration of the optimal solution. The proposed strategy to deal with this dynamic context is to build an incremental process starting with a local suboptimal solution and then refine it gradually as new regions in the search space are observable.

Ideally, the robot must avoid revisiting the same area twice to optimize the exploration time and reduce energy consumption, but this is not always possible. An example is when the robot is trapped in a dead end or a closed path and must turn back to find an alternative way. Thus, revisiting the same area is not forbidden such as in the traveling salesman problem, but it needs to be minimized.

In this paper, we use the Butterfly Optimization Algorithm (BOA) to solve the unknown area exploration problem with energy constraints in dynamic environments.

The motivation behind this work is the fact that BOA got promising results in other global optimization problems and was never used for solving robotics problems before, as far as we know. We also propose a new version of this algorithm called xBOA based on the crossover operator to improve the diversity of the candidate solutions and speed up the convergence of the algorithm.

In parallel with this contribution, we propose a framework for solving the unknown area exploration problem with energy constraints in both single- and multi-robot scenarios using metaheuristics. The primary goal of this framework is to create a benchmarking suite adapted to dynamic incremental problems in robotics such as exploration tasks. Thus, the framework is made in such a manner to be generic to easily compare different metaheuristics with minimum modifications.

We used this framework to evaluate the performances of the newly proposed algorithm xBOA and compare it with 5 other metaheuristics widely used in the literature through different evaluation criteria. We also compared the results of xBOA with the original version of BOA and 3 other variants recently introduced in the literature: SABOA, mBOA, and ABOA. Finally, we validated the adaptability of our approach to multi-robot scenarios.

The evaluation criteria used for this work are as follows: (1) the percentage of the exploration area reached by the robots given a certain amount of energy, (2) the time needed by the robots to finish the exploration mission, (3) the execution time of the metaheuristic, (4) the convergence speed of the metaheuristic, and (5) the number of fitness evaluations required by the metaheuristic.

To summarize, the contributions presented in this paper are:

- Propose a new version of the Butterfly Optimization Algorithm (named xBOA) based on the crossover operator. And compare its results to well-known metaheuristics used in similar works, as well as the original BOA, and 3 other variants recently introduced in the literature.
- Propose a framework for benchmarking different techniques used to solve the unknown area exploration problem with energy constraints using metaheuristics in both single- and multi-robot scenarios. The framework is made in such a manner to be generic to easily compare different metaheuristics with minimum modifications.
- Compare the performance of xBOA with well-known metaheuristics used in the literature using 5 different comparison criteria.
- Propose an adapted implementation of BOA and its other variants to the Pygmo2 library.

In the following section, we present the state of the art of robotics exploration. Section 3 describes the Butterfly Optimization Algorithm and the proposed variant. We present our modelization and methodology in Sect. 4 and the experimental results in Sect. 5.

## 2 Related work

Many techniques have been used in robotics for area exploration. These techniques can be classified according to several criteria regarding their determinism, the necessity to use prior information, or the adaptability to a multi-robot context. The exploration problem consists of discovering an area to gather information Li (2020). A variant consists of visiting every possible point of the area; it is known as the complete coverage problem Li (2020) and is mainly used for cleaning robots. Another variant consists of continuously exploring an area to detect potential intruders; this is known in the literature as the patrolling problem or consistent coverage Hoshino and Takahashi (2019).

A popular deterministic method for solving the unknown area exploration problem has been introduced by Yamauchi (1997) where the robot keeps moving toward the nearest frontier point. Frontiers are the extremity lines separating explored and unexplored regions. This technique is easy to implement and requires few computational resources, yet it gives good results in practice. Many variants of the same algorithm appear in the literature trying to find the best strategy for selecting the next frontier point Holz et al. (2010). The original authors also published the multi-robot version of this algorithm Yamauchi (1998) where each robot moves toward its nearest frontier; however, this strategy fails to avoid redundancy since several robots can be assigned to the same frontier. Bautin et al. (2012) tried resolving the problem by using wavefront propagation to spread the robots. The frontiers are assigned a rank according to how many robots are close to them. Each robot is then assigned to a different frontier that has a small rank, which allows it to get as far as possible from the other robots and maximizes the exploration area.

The authors of Al khawaldah and Nuchter (2015) choose a different strategy for indoor environments: the first robot explores the entire corridor and detects the doorways; then, each one of the other robots selects a different doorway and explores the corresponding room using the frontier-based technique. The robot does not exit a room until it explores it entirely. This strategy encourages the robots to explore the rooms individually and reduce the overlapping between their assigned regions, which contributes to reducing the total mission time. More recently, Luperto et al (2020) use incomplete data such as simplified evacuation maps or floor plans as an input for a cost function to choose the best frontier to explore. The experiments show that exploiting approximative maps can speed up the exploration mission, even if the data are inaccurate. However, it requires the maps to be manually cleaned and aligned by a human operator.

Another family of deterministic approaches relies on decomposing the environment into several sub-regions and then exploring each region independently using a simple strategy such as a back-and-forth or a circular shape motion. One popular technique from this family is the Boustrophedon decomposition Choset and Pignon (1998). It decomposes the map into polygonal regions based on the position of obstacles. It was successfully used for complete coverage tasks. However, it assumes a static environment (i.e., no moving obstacles) and requires the map to be known in advance. Other techniques use Voronoi diagrams to decompose the map into more flexible regions García et al (2007); Masehian and Amin-Naseri (2004).

Dakulović et al (2011) used wavefront propagation to adapt the D* algorithm to the coverage problem. Instead of planning a path from a starting position toward a goal location, the modified D* successfully generated a path to visit all the points of a given map. The fast-replanning feature of the D* algorithm allows it to adapt to dynamic environments by quickly modifying the path in case an obstacle changes its position.

Song and Gupta (2018) proposed a new approach for generating effective coverage paths. It uses a multilayer maps representation called Exploratory Turing Machine to produce a back-and-forth trajectory with an adjustable sweep direction. This strategy results in shorter trajectory lengths compared to classical methods based on back-and-forth motions. Shen et al (2020) extended it recently by adding energy constraints. The robot executes the coverage path until its energy is low; then, it returns to the charging station to refill its battery. After that, it restarts the coverage at a nearby unexplored region to avoid a long travel distance to the previous point when it stopped. This ensures the complete coverage of the environment with a reduced overlap.

Learning-based methods have also been used for the robotics exploration problem. In the approach proposed by Tai and Liu (2016), the robot does not rely on any map to explore the environment. It uses an end-to-end Deep Q-Network to choose the most suitable direction to follow using only camera images as input. It has been tested to navigate inside an unknown corridor environment while avoiding the walls. On the other hand, Ström et al (2017) relies on a database of previously seen maps to predict unknown regions of a partially explored map. It uses a bag-of-words inspired technique to detect similarities between grid maps and learn to complete the missing areas. This helps plan paths beyond the explored region, which reduced the distance traveled by the robot compared to the frontier-based exploration method. Similarly, Shrestha et al (2019) predicts unknown regions beyond frontiers using Variational Autoencoders and then a cost-utility heuristic to choose which one to explore next.

In the category of stochastic methods, metaheuristics have been widely used in different areas of robotics Fong et al (2015) and are still widely used for both ground and aerial robots. Ahmadi et al (2018) used a Genetic Algorithm (GA) to monitor a known area using an aerial robot, while satisfying some constraints such as the path's length and smoothness. Zhou et al (2013) used the Particle Swarm Optimization (PSO) algorithm with the frontier-based strategy to optimize the effectiveness of the exploration task. Xiao et al (2013) used Ant Colony Optimization (ACO) for multi-robot exploration. Recently, Kamalova et al (2020) used Grey Wolf Optimizer (GWO) to select the next frontier point to explore. They also proposed a multi-objective version of the same algorithm to maximize the explored area and map accuracy Kamalova et al (2019).

Although the most used metaheuristics in the robotics field are generally the classic techniques used to solve global optimization problems, other metaheuristics can also be applied regarding the No-Free-Lunch theorem Wolpert and Macready (1997) which states that no algorithm is better than another algorithm in all types of problems. That means if one technique shows superior results in some classes of problems, it cannot show superior results for all other classes.

This theorem has motivated researchers to invent new metaheuristics and apply them to different fields, including robotics. However, there are many new metaheuristics that have not yet been used in the context of area exploration. Some examples of these recently developed techniques include, as far as we know: the Butterfly Optimization Algorithm (BOA) Arora and Singh (2019), Atomic Orbital Search Azizi (2021), Dwarf Mongoose Optimization Algorithm Agushaka et al (2022), Arithmetic Optimization Algorithm Abualigah et al (2021a), Tuna Swarm Optimization Xie et al (2021), Aquila optimizer Abualigah et al (2021b), and Reptile Search Algorithm Abualigah et al (2022).

We present a comparative summary between the cited approaches in Table 1.

# 3 Butterfly Optimization Algorithm

## 3.1 Theory and biological inspiration

Butterfly Optimization Algorithm (BOA) is a recent metaheuristic introduced by Aurora et al. Arora and Singh (2019). It is inspired from the foraging/mating behavior of butterflies.

At the beginning of the algorithm, a population of random solutions (i.e., butterflies) is generated and refined incrementally through several iterations until a certain stopping condition is met.

In nature, butterflies rely on smell for finding food sources and mating partners. They use sensors in their bodies to perceive these smells (fragrance) and measure their intensity Arora and Singh (2019). The more intense the fragrance is, the more attractive is the butterfly toward the source of this fragrance.

The BOA algorithm models this behavior by computing a fragrance value proportional to the fitness of the individual. The better is the fitness, the bigger is the fragrance, and thus, the most attractive is the butterfly. In other terms, butterflies will move gradually toward the best butterfly in the search space. This is known as the global search phase Arora and Singh (2019).

In order to avoid premature convergence, a local search phase is executed in which the butterflies move randomly in the local region where they are located. The alternation between the global search and local search phases is controlled by using a probability parameter called the switching probability.

In nature, the fragrance emitted by a butterfly might be altered by weather conditions; two parameters are then introduced in the algorithm to modify the intensity of the fragrance that will be sensed by the other butterflies. These parameters are the sensor modality and power exponent. Equation 1 describes this operation:

**Table 1** Comparative summary between the cited approaches

| References | Map | Approach family | Approach type | Energy limit. | Experiment | Number of robots | Exploration type |
|---|---|---|---|---|---|---|---|
| Yamauchi (1997) | Unknown | Frontier-based | Deterministic | No | Real robot | Single | Exploration |
| Al khawaldah and Nuchter (2015) | Unknown | Frontier-based | Deterministic | No | Simulation | Multi-robot | Exploration |
| Bautin et al. (2012) | Unknown | Wavefront propagat. | Deterministic | No | Simulation | Multi-robot | Exploration |
| Luperto et al (2020) | Partially known | Frontier-based | Deterministic | No | Simulation & real robot | Single | Exploration |
| Dakulović et al (2011) | Known | D* | Deterministic | No | Simulation | Single | Complete coverage |
| Song and Gupta (2018) | Known | $\epsilon$* | Deterministic | No | Simulation & real robot | Single | Complete coverage |
| Shen et al (2020) | Unknown | $\epsilon$* | Deterministic | Yes | Simulation | Single | Complete coverage |
| Ahmadi et al (2018) | Known | GA | Stochastic | Yes | Simulation | Single | Exploration |
| Kamalova et al (2020) | Unknown | GWO | Stochastic | No | Simulation & real robot | Single | Exploration |
| Kamalova et al (2019) | Unknown | GWO | Stochastic | No | Simulation | Multi-robot | Exploration |
| Tai and Liu (2016) | – | Deep Q-Network | Learning-based | No | Simulation | Single | Exploration |
| Ström et al (2017) | Unknown | FabMap2 | Learning-based | No | Real robot | Single | Exploration |
| Shrestha et al (2019) | Partially known | Variational autoencod. | Learning-based | No | Simulation | Single | Exploration |
| Our approach | Unknown | BOA | Stochastic | Yes | Simulation | Single & multi-robot | Exploration |

$$F = c * I^a \quad (1)$$

where $I$ is the intensity, modelized by the fitness value of the butterfly, $c$ is the sensor modality, and $a$ is the power exponent.

The following equations describe how individuals are updated during the global and local search phase, respectively:

$$x_i^{t+1} = x_i^t + (r^2 * g^* - x_i^t) * f_i \quad (2)$$

$$x_i^{t+1} = x_i^t + (r^2 * x_j^t - x_k^t) * f_i \quad (3)$$

where

- $x_i$ is the current butterfly
- $r$ is a random number in the interval [0,1]
- $g^*$ is the best butterfly in the population
- $f_i$ is the fragrance of the $i^{th}$ butterfly
- $x_j$ and $x_k$ are two butterflies picked randomly from the population

In the BOA algorithm, each butterfly moves in the search space according to Eqs. 2 and 3, which update the fitness value of the butterfly. Then, we compute the best solution having the best fitness value. When this process is repeated, the butterflies will converge toward the optimal solution in our search space. However, a premature convergence may lead to being trapped in a local optimum, which is why the butterflies will move randomly during the local search phase to explore new potential solutions.

Another important step added by the authors consists of updating the value of the sensor modality ($c$) parameter at each iteration Arora and Singh (2016). The goal is to avoid premature convergence that might happen if we set a value of $c$ too big or too small. The following rule was used by the authors and showed good results compared to the classical BOA:

$$c^{t+1} = c^t + \left( \frac{0.025}{c^t * \text{max_iterations}} \right) \quad (4)$$

Algorithm 1 describes the pseudo-code of the BOA algorithm.

**Algorithm 1** Butterfly Optimization Algorithm (BOA)

```
Initialize sensor modality (c), power exponent (a), and switch probability (p)
Define the fitness function:
  f(x) = (x_1, x_2..., x_dim)
Generate the initial population of n butterflies
while stopping criteria not met do
  for each butterfly bf in population do
    Calculate fragrance for bf using:
    F ⇐ c * I^a
  end for
  Find the best bf
  for each butterfly bf in population do
    Generate a random number r ∈ [0, 1]
    if r < p then
      Move toward best butterfly g*:
      x_i^{t+1} ⇐ x_i^t + (r^2 * g* − x_i^t) * f_i
    else
      Move randomly (local search):
      x_i^{t+1} ⇐ x_i^t + (r^2 * x_j^t − x_k^t) * f_i
    end if
    Evaluate the new solution
    Replace the old solution if the new solution is better
  end for
  Update the value of c
end while
```

### 3.2 xBOA algorithm

We propose a modified version of BOA based on the crossover operator inspired from the genetic algorithm Goldberg (1989). The intuition behind this proposition is the fact that Eq. 2 moves all individuals toward the current best solution of the population, ignoring other candidate solutions having the same fitness value or having the potential to become better individuals after few iterations.

To avoid this problem, we replace Eq. 2 with the crossover operator during the global search phase.

Several combination strategies have been introduced in the literature [ Pavai and Geetha (2016). For simplicity purposes, we use the single-point crossover strategy. It consists in cutting a parent individual into two sub-vectors and then swapping them with the second parent's sub-vectors to produce two new individuals (see Algorithm 2).

The following example shows the output of this operation for two individuals of size 5:

$$\text{Parent1} = [\boldsymbol{x_1}, \boldsymbol{x_2} \| \boldsymbol{x_3}, \boldsymbol{x_4}, \boldsymbol{x_5}]$$
$$\text{Parent2} = [y_1, y_2 \| y_3, y_4, y_5]$$
$$\text{Offspring1} = [\boldsymbol{x_1}, \boldsymbol{x_2}, y_3, y_4, y_5]$$
$$\text{Offspring2} = [y_1, y_2, \boldsymbol{x_3}, \boldsymbol{x_4}, \boldsymbol{x_5}]$$

**Algorithm 2** Crossover operator

```
Select a random partner x_j, where j ≠ i
Cut x_i and x_j in two parts:
  x^i = [x_1^i, x_2^i]
  x^j = [x_1^j, x_2^j]
Swap and combine to produce new offspring:
  child_1 ⇐ [x_1^i, x_2^j]
  child_2 ⇐ [x_1^j, x_2^i]
```

Creating new individuals might make the population grow exponentially. Thus, instead of inserting the new offspring into the population directly, we remove the parent and replace

**Algorithm 3** Crossover Butterfly Optimization Algorithm (xBOA)

Initialize sensor modality ($c$), power exponent ($a$), and crossover probability ($p$)
Define the fitness function:
  $f(x) = (x_1, x_2..., x_{dim})$
Generate the initial population of $n$ butterflies
**while** stopping criteria not met **do**
  **for** each butterfly $bf$ in population **do**
    Calculate fragrance for $bf$ using:
    $F \Leftarrow c * I^a$
  **end for**
  Find the best $bf$
  **for** each butterfly $bf$ in population **do**
    Generate a random number $r \in [0, 1]$
    **if** $r < p$ **then**
      Select a random partner $x^j$
      Generate two child individuals using the crossover operator
      Evaluate the new individuals
      Replace parent by the best child if it has a better fitness value
    **else**
      Move randomly:
      $x_i^{t+1} \Leftarrow x_i^t + (r^2 * x_j^t - x_k^t) * f_i$
    **end if**
  **end for**
  Update the value of $c$
**end while**

it with its best offspring. This allows us to keep a fixed population size during the entire process.

By using the crossover operator, we encourage the butterflies to move toward several candidate solutions instead of converging only toward the best known solution. This creates more diversity in the population and encourages the algorithm to escape the trap of premature convergence toward a local optimum. In other words, the population of butterflies will investigate several regions of the search space simultaneously in order to find the global best solution faster.

However, we need to ensure keeping a good tradeoff between the exploration and exploitation features of the algorithm by balancing the global and local search phases. To do that, we only use the crossover operator if a certain probability is met. The switch probability parameter becomes then the crossover probability.

Another key difference with the original BOA resides in the local search phase. In xBOA, the equation 3 is applied even if it decreases the quality of the solution. This seems counterproductive in the short term, but it allows the algorithm to increase the diversity of the solutions by allowing them to move randomly in the search space and explore new regions that might contain the global optimum. In other words, it is sometimes advantageous to lose the quality of an individual to unlock better solutions in the next generations.

The experiment results in Sect. 5 showed that the proposed modifications make xBOA more robust to local optima than the original BOA and require less iterations to find the best solution.

Algorithm 3 describes all the operations of xBOA.

### 3.3 Other variants of BOA

Although BOA is a recent metaheuristic, it has many variants and has been used successfully to solve several problems. The first version of this method was proposed by Arora and Singh (2015) in a conference paper with a code logic similar to Flower Pollination Algorithm Yang (2012) and using a Lévy Flight distribution for the updating rules, in later papers they changed the equations and added a rule to automatically update the value of the sensor modality, which gave better convergence rates Arora and Singh (2016, 2019); this last version was proposed as the official BOA algorithm. In another paper, they proposed a binary variant of the method for the features selection problematic Arora and Anand (2019).

Other authors proposed the hybridization of the algorithm with other methods. Jalali et al (2019) used a hybrid BOA-MLP (Multi-Layer Perceptron) for data classification. The MLP helped to keep a good balance between the exploration and exploitation phases of BOA and improved the results. Zhang et al. (2020) created a hybrid PSO-BOA approach, coupled with the chaotic theory to get better results for high-dimensional problems. The authors also proposed a new non-linear updating rule for the power exponent parameter which shows better results than the classic linear rule. Assiri (2021) also used the chaotic theory as a local search phase and applied it to solve a feature selection problem. Wang et al (2021) used a hybridization with Flower Pollination Algorithm-based on mutualism mechanism; the technique gives good results but suffers from a long running time. The mutualism principle was also used by Sharma et al (2021) for hybridization with the Symbiosis Organisms Search. Zounemat-Kermani et al (2021) combined the algorithm with the Adaptive Neuro-Fuzzy Inference System (ANFIS) and got better performances than the classic ANFIS and the hybrid ANFIS-FireFly Algorithm.

Another direction toward creating new BOA variants focuses on improving the algorithm's logic. The authors of the method proposed a modified version called mBOA Arora et al (2018) that includes a new step after the local/global search phases. The goal is to add an intensive exploitation search to avoid getting trapped in local optima. The results showed faster convergence compared to the original approach. Tubishat et al. (2020) included a local search using the mutation operator to improve the performance of the method. The approach outperformed the other methods used by the authors in 15 of 20 benchmark datasets, but it also suffers from a long execution time. Li et al (2019) used the cross-entropy method and a co-evolution technique to successfully solve 3 classical engineering problems. Fan et al. (2020) opted for reducing the number of hyperparameters by introducing a new self-adaptative rule for the fragrance formula and eliminating the sensor modality and power exponent coefficients. The new variant is named SABOA (Self-Adaptative BOA). On the other hand, Guo et al (2021) used a guiding weight factor in the global search equation for improving the convergence speed and a population restart strategy to avoid getting trapped into local minima. The results of these new variants showed promising results toward solving high-dimensional optimization problems, which is one of the areas where the classical BOA does not perform well.

## 4 Modelization and methodology

### 4.1 Occupancy grid maps

Unknown area exploration is closely related to the navigation and mapping problem. The robot has to move in the environment and discover it gradually. During this operation, it is highly probable to meet dead ends and other obstacles blocking the way. The robot will then memorize the position of obstacles and use them to plan alternative paths and discover new regions.

Robots use sensors to detect walls and obstacles. Since these sensors have a limited range, it is not possible to observe the entire environment at once. In this case, we need to store the positions of the detected obstacles inside a data structure that allows the robot to easily aggregate new observations and combine them in a way that simplifies trajectories calculation.

The Occupancy Grid Map Elfes (1989) is the most used data structure for representing the environment in robotics. It is a 2D matrix where each cell represents a portion of the environment. The size of the cells influences the degree of details shown on the grid; it is usually set to a size equal to or lower than the robot footprint. Figure 1 shows an example of a grid map.

The value of each cell in the grid represents the probability that the corresponding region in the environment is empty or occupied by an obstacle. Since the robot has no prior information about the region to explore, all the cells have a prior probability of occupation of 0.5, which will be updated using the Bayes rule (Eq. 5) each time the robot's sensors observe the corresponding region.

$$p(A/B) = \frac{p(B/A) * p(A)}{p(B)} \tag{5}$$

A is the occupancy value, and

B is the observation

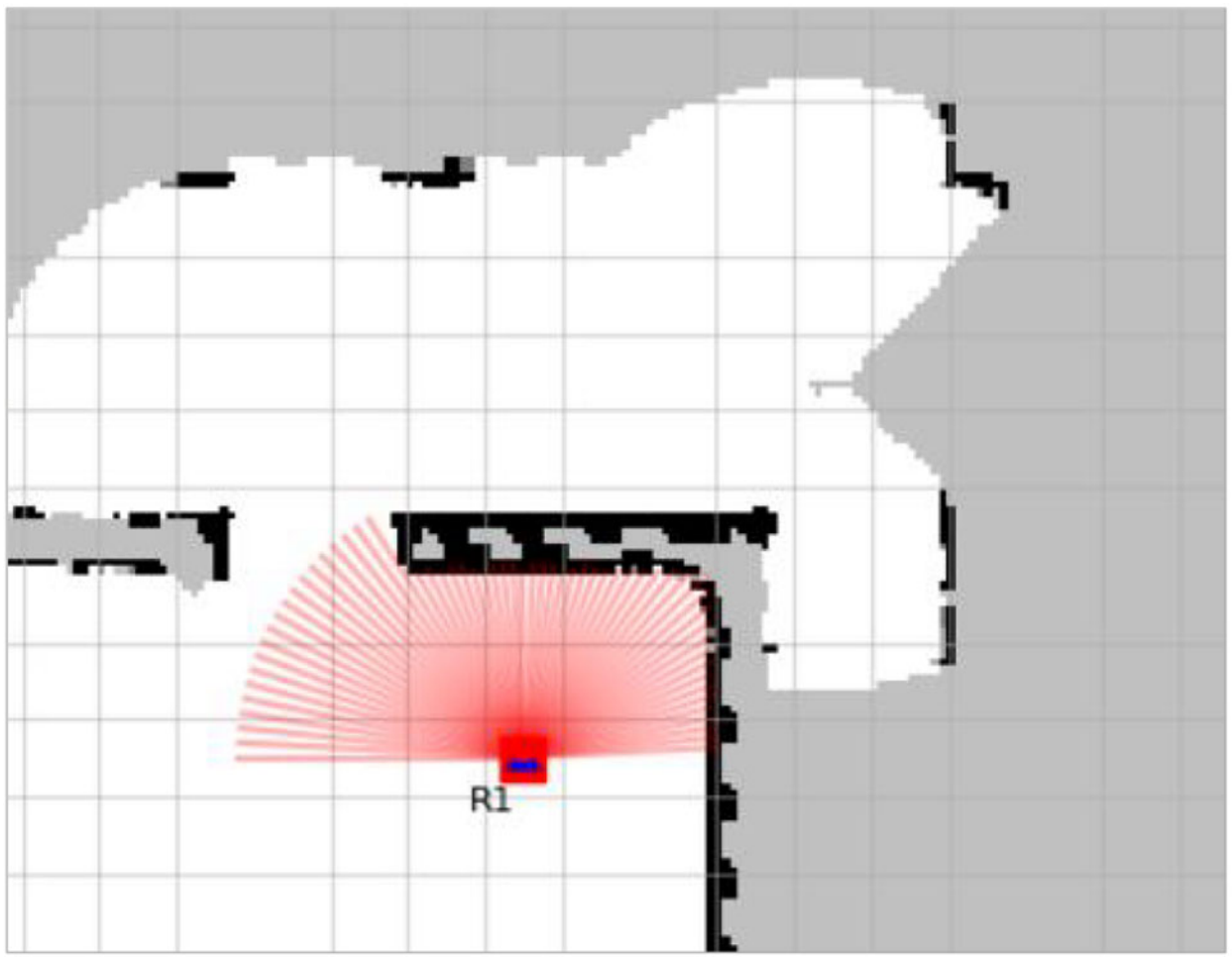

**Fig. 1** An example of occupancy grid maps. White pixels represent the explored area; Gray pixels represent the unknown area; Black pixels represent the detected obstacles; Red lines represent the Laser sensor beams

It is a frequent practice to use the log odds representation instead of probabilities to convert the multiplications operations into additions as in the following equations:

$$\begin{aligned} \text{odds(A)} &= \frac{p(A)}{P(\neg A)} \\ \text{odds(A/B)} &= \frac{p(A/B)}{P(\neg A/B)} \end{aligned} \tag{6}$$

After applying Eq. 6 into the Bayes rule, we get Eq. 7. The log odds value vary between $[-\infty,+\infty]$, which is useful for avoiding the multiplication of small numbers during the implementation that may cause some issues because of the limited precision of float values.

$$\text{logodds(A/B)} = \log\frac{\text{p(B/A)}}{\text{P(B/}\neg\text{A)}} + \text{logodds(A)} \tag{7}$$

As a result, we can label each cell $C_{ij}$ in 3 possible categories:

$$C_{ij} \text{ is } \begin{cases} Occupied\ if\ Occ(C_{ij}) > 0 \\ Empty\ \ if\ Occ(C_{ij}) < 0 \\ Unknown\ if\ Occ(C_{ij}) = 0 \end{cases} \tag{8}$$

where $Occ(C_{ij})$ is the occupancy log odds from Eq. 7. We use 0 as a threshold value because $logodds(0.5) = 0$.

### 4.2 Unknown area exploration using xBOA

Unknown area exploration is often modeled as an optimization problem; the goal of this process is to assign an occupancy probability to every cell in the map. In order to achieve this goal, the robot has to maximize the surface of the explored area while minimizing energy used.

The role of the xBOA metaheuristic is essential in this process, it starts by generating a population of random target locations to visit—which will be the butterflies, and then enhances the position of these target locations through a succession of local search and crossover operations.

Mathematically, each candidate solution $X_k$ in the population represents a set of target cell locations $C_{ij}$, where $(i, j)$ are $(x, y)$-coordinates inside the grid map boundaries.

$$X = C_{ij}$$

Therefore, the fitness function can be modeled as a maximization of the number of newly observed cells which have an occupancy log odds value equal to 0 (a.k.a. unexplored cells). Equation 9 defines the mathematical formulation of this function.

$$F = \max(\text{Observed Cells}) = \min\left(\sum_{i,j}(\delta(C_{ij,0}))\right)$$
$$\text{where } \delta(C_{ij,0}) = \begin{cases} 1 \; if \; Occ(C_{ij}) \neq 0 \\ 0 \quad \text{otherwise} \end{cases} \quad (9)$$

With the following constraint:

$$\sum_{i,j} E(C_{ij}) < \text{current battery level}$$

where $E(C_{ij})$ is the energy needed for moving the robot from the actual position to cell $C_{ij}$.

Once the best set of target locations that satisfies the energy constraint is found, the robot calculates the shortest path that links these target locations using the $A^*$ algorithm Hart et al (1968); then, it executes this path until visiting all target locations. After that, it repeats the optimization algorithm to generate a new set of target locations and continues the process until all the cells in the map have been observed (i.e., the robot explored the entire area).

It is important to recall that the planned path is not necessarily optimal, since the robot cannot detect obstacles outside the range of its sensors. Moreover, it has no prior information about the region to explore. Thus, it is highly probable to meet dead ends and other obstacles blocking the way during the navigation. The robot is then forced to find an alternative path to escape the impasse, even if this requires returning to a previously visited region and consume more energy.

Consequently, the solution is built incrementally, starting with a sub-optimal path and then updating it regularly as new obstacles are observed, which also ensures that the robot adapts to dynamic environments and avoids moving obstacles.

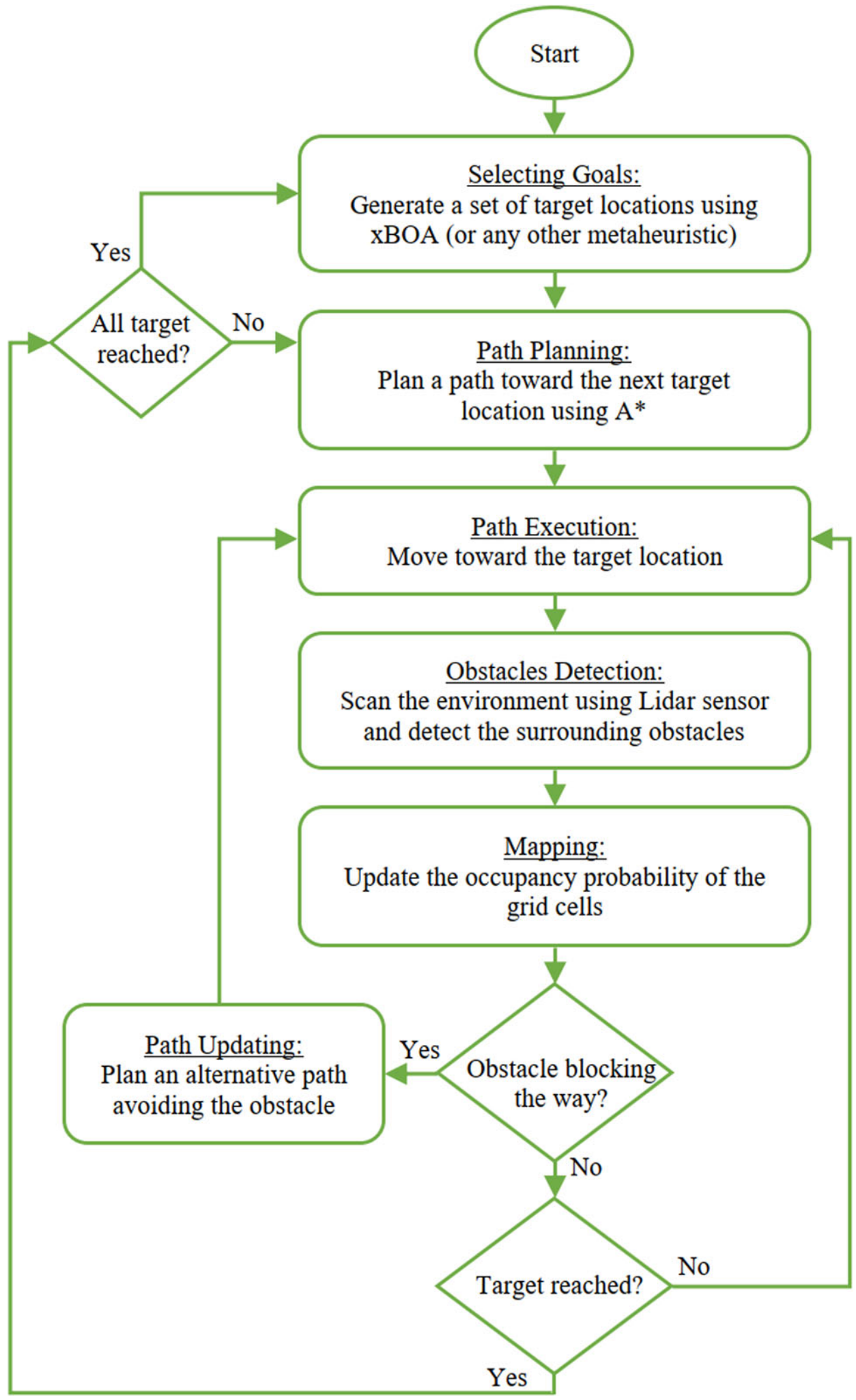

**Fig. 2** Workflow of the unknown area exploration process

Figure 2 summarizes the general workflow of this process.

Using Eq. 9 as fitness function means that for each candidate solution in the population, we will plan a path toward the target locations defined by this solution and then estimate how many new cells will be observed if this path is executed by the robot. The complexity of this operation is $O(M^2)$ where M is the path length.

Since the fitness value is evaluated for each candidate solution in each generation, the overall complexity becomes $O(N \times K \times M^2)$ where N is the population size, K is the number of generations, and M is the path length. Figure 3 illustrates the process of fitness evaluation for a population of 4 candidate solutions.

If the population size or the number of generations is set to a big value, the process becomes computationally expensive. An alternative approach would be to compute a rough estimate of the observed cells (using a distance vector for

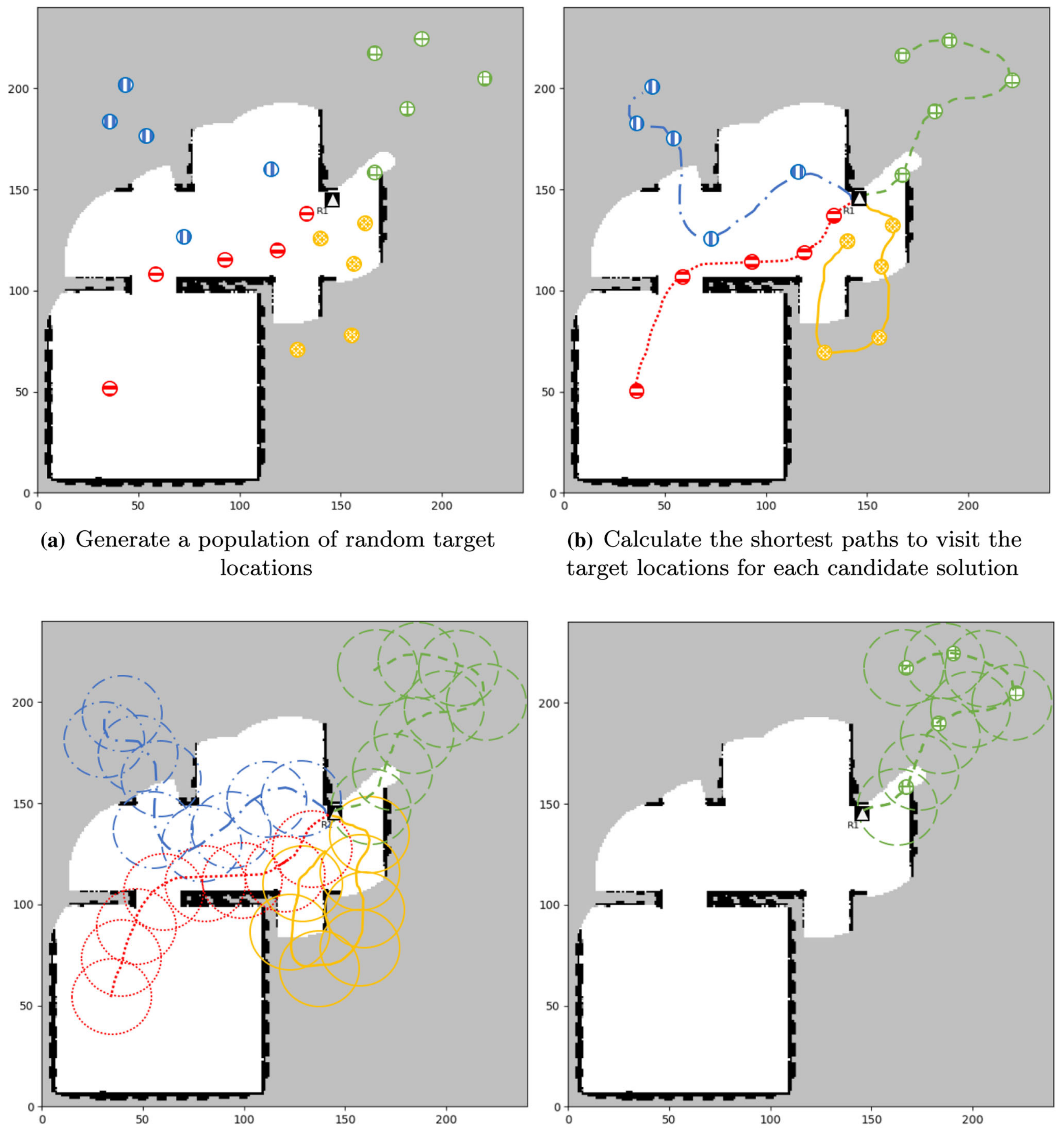

(a) Generate a population of random target locations

(b) Calculate the shortest paths to visit the target locations for each candidate solution

(c) Estimate the surface of the newly explored area for each path

(d) The best solution is the one maximizing the surface of the newly explored area

**Fig. 3** An example of the fitness evaluation process for a population of 4 candidate solutions

example) as a fitness criterion, but that would decrease the quality of the generated solutions. Neither strategy is perfect. In our situation, we choose to apply the first strategy while reducing the population size to get an acceptable tradeoff between the quality of the solutions and the execution time. In case the algorithm has to be executed in a machine with limited CPU resources (such as robots with small onboard computers), we would suggest using the second solution.

## 5 Experiments and analysis

### 5.1 Simulation environment

A common practice in robotics research is to first validate the approach in a simulation environment before switching to real-world experiments. This allows us to safely validate a method and avoid material damage in case of errors in the code, while it gives the possibility of repeating an experiment a certain number of times in the same conditions.

There are several popular 3D simulators for robotics experimentation in the literature Dosovitskiy et al (2017); Koenig and Howard (2004); Rohmer et al (2013). Those simulators are highly realistic but require heavy computations and a powerful machine. Since our goal is to test the effectiveness of the metaheuristics, we do not need all the features provided by those kinds of simulators; consequently, we implemented a custom simulator dedicated to benchmark the robotics exploration techniques.

We use python for its popularity in the robotics field, its portability to different platforms, and a large number of libraries available for performing scientific computations, although it presents some limitations in the threading model because of the Global Interpreter Lock implementation Beazley (2010).

To proceed successfully with our experiments, we implemented a navigation model for holonomic robots based on the most common robotics platforms used in research. The robot observes it's surrounding environment using a 180° Laser Range sensor (LIDAR) for calculating the distance with close objects (see Figure 1). The LIDAR has a 95% precision rate, which means that the observations are subject to errors because of the presence of non-perfect conditions in the real world. We simulate imprecision by adding Gaussian noise to the measurements.

The robot can move in 8 possible directions, at each move it turns toward the target location, then moves forward. During the rotations, the robot continues scanning the environment using its sensors to locate new obstacles. In the case it detects an obstacle blocking its way, the robot stops and then computes an alternative path based on the new information registered in the occupancy grid map.

### 5.2 Experiments setup

For evaluating the efficiency of the Butterfly Optimization Algorithm, we compare it to several metaheuristics widely used in the literature:

- Artificial Bees Colony (ABC); Karaboga (2005)
- Covariance Matrix Analysis Evolution Strategy (CMAES); Hansen et al (2003)
- Genetic Algorithm (GA); Goldberg (1989)
- Grey Wolf Optimizer (GWO); Mirjalili et al (2014)
- Particle Swarm Optimization (PSO); Kennedy and Eberhart (1995)

**Table 2** Best hyperparameters found after 30 execution trials

| Methods | Hyperparameters | Values | Methods | Hyperparameters | Values |
|---|---|---|---|---|---|
| BOA (pop size 20) | Power exponent | 0.547 | BOA (pop size 5) | Power exponent | 0.73 |
| | Sensor modality | 0.602 | | Sensor modality | 0.577 |
| | Switch probability | 0.395 | | Switch probability | 0.331 |
| xBOA (pop size 20) | Power exponent | 0.905 | xBOA (pop size 5) | Power exponent | 0.994 |
| | Sensor modality | 0.257 | | Sensor modality | 0.518 |
| | Crossover probability | 0.593 | | Crossover probability | 0.583 |
| GA | Crossover probability | 0.11 | mBOA (pop size 5) | Power exponent | 0.61 |
| | Mutation probability | 0.215 | | Sensor modality | 0.356 |
| | Mutation distribution index | 76.026 | | Switch probability | 0.762 |
| PSO | Social component coef. | 1.506 | ABOA (pop size 5) | Power exponent | 0.992 |
| | Cognitive component coef. | 3.379 | | Sensor modality | 0.98 |
| | Max velocity | 0.329 | | Switch probability | 0.983 |
| | Inertia weight | 0.449 | | $\mu$ | 1.356 |
| ABC, GWO, CMAES | No parameters to optimize (auto-tuning) | | SABOA (pop size 5) | Switch probability | 0.237 |

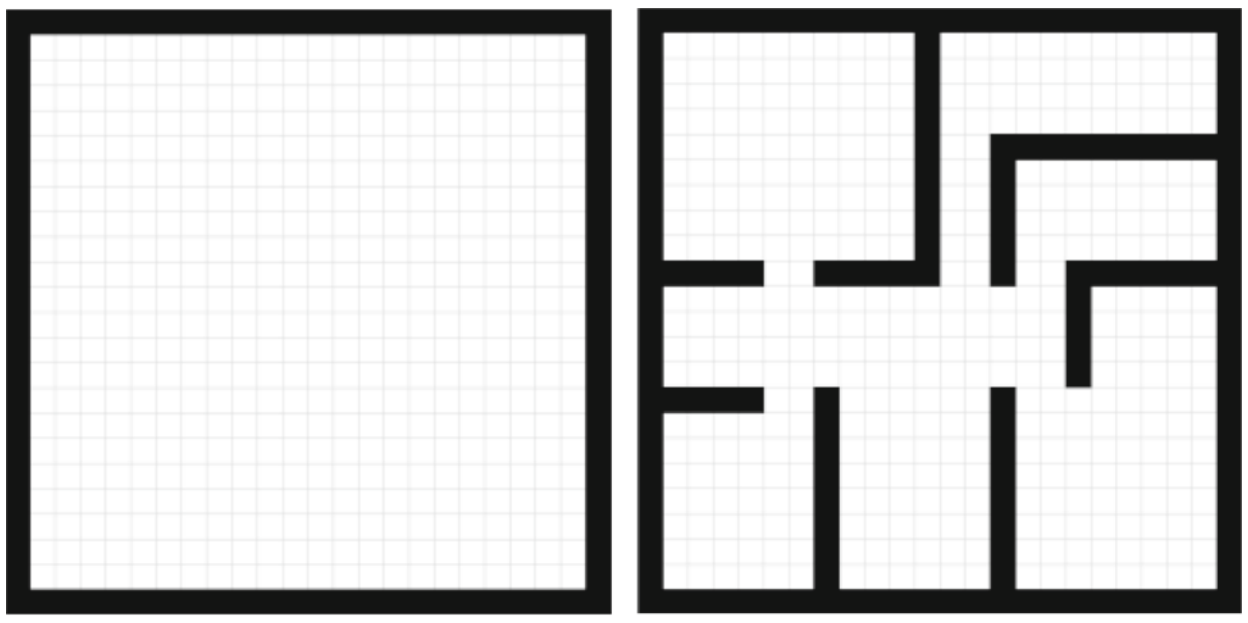

(a) Empty map (24x24m)

(b) House map (24x24m)

**Fig. 4** Maps used during the experiments

Each one of these metaheuristics is used as an optimizer during the exploration process. Except for BOA and xBOA, we used the implementations provided by Pygmo2 Biscani and Izzo (2020), which is a python library aiming to offer a unified interface for implementing massively parallel optimization algorithms.

Since these methods are based on stochastic operators, we feed the same initial population to each optimizer and repeat the execution 10 times. We set a stopping condition when the energy of the robot reaches 0, or when the percentage of the explored area reaches a rate bigger than 99%.

We conducted another set of experiments to evaluate the performance of xBOA compared to other variants of BOA cited in the following list:

- Self-adaptative BOA (SABOA); Fan et al. (2020)
- BOA with intensive search (mBOA); Arora et al (2018)
- BOA with nonlinear adaptative rule (ABOA); Zhang et al. (2020)

We reimplemented BOA and each one of these variants in Python, and adapted them to the Pygmo2 library, to ensure the difference in the results is not caused by different implementations techniques or different libraries.

Finally, we run a test to validate the adaptability of our approach to multi-robot scenarios.

Figure 4 shows different environments used in the experiments. The first one is an empty map with no obstacles except the surrounding perimeter wall. The second one is a moderately occupied scene inspired by the architecture of a house with an obstacle presence rate of 27%. These two maps have a size of 24 × 24 m. Note that the doors were explicitly removed from the maps for simplifying the experiments.

For each of these maps, we conducted two types of experiments varying the number of target goals generated by the optimizer. This parameter influences the strategy of exploration: a small number of goals will make the robot plan for short-term exploration, while a bigger number will make it plan for long-term exploration. For each strategy, we conducted a set of experiments to analyze the execution time and the performance of the optimizers. These tests have been executed using an i7 2.8Ghz CPU laptop. Each test has been repeated 10 times for each optimizer.

To reduce experiments variability, we set the same starting conditions for all sessions:

- Robot starting location at position (1,1) with angle 0° to the north.
- LIDAR distance range: 4m, with a 5% error rate.
- LIDAR angular range: 180°, with 1° resolution.
- Population size: 20.
- Max number of generations: 30.
- Early stopping: if no improvement in the fitness value during 10 consecutive generations.
- Initial population: same for all methods.
- The seed for random numbers generation: 25.

We used the following evaluation measures to compare the performance of the optimizers:

- Step number: Number of moves and rotations performed by the robot. It is proportional to the amount of energy consumed.
- Execution time: Total duration—in seconds—since the beginning of the mission.
- Exploration rate: Surface of the area observed by the robot sensors and added to the map.
- The number of fitness evaluations: Number of candidate solutions evaluated using the fitness function.
- The computation time: Amount of time required by the metaheuristic to execute all the iterations and return the best solution.

## 5.3 Hyperparameters search

Metaheuristics are known to be sensitive to parameters initialization. It is frequent in the literature to compare the performance of the methods using the original parameters published by the first authors Arora and Singh (2019). This may be a good strategy if we are comparing them using standard benchmarking functions; however, the robot exploration problem is by nature an incremental partly observable dynamic problem, which makes the standard parameters values not necessary the best possible values.

Thus, we conducted a preliminary experiment to search for the best hyperparameters for each method according to this problem. This will allow us to compare these metaheuristics in their best possible performance and avoid some sensitive methods being trapped in a local optimum because of bad initializations.

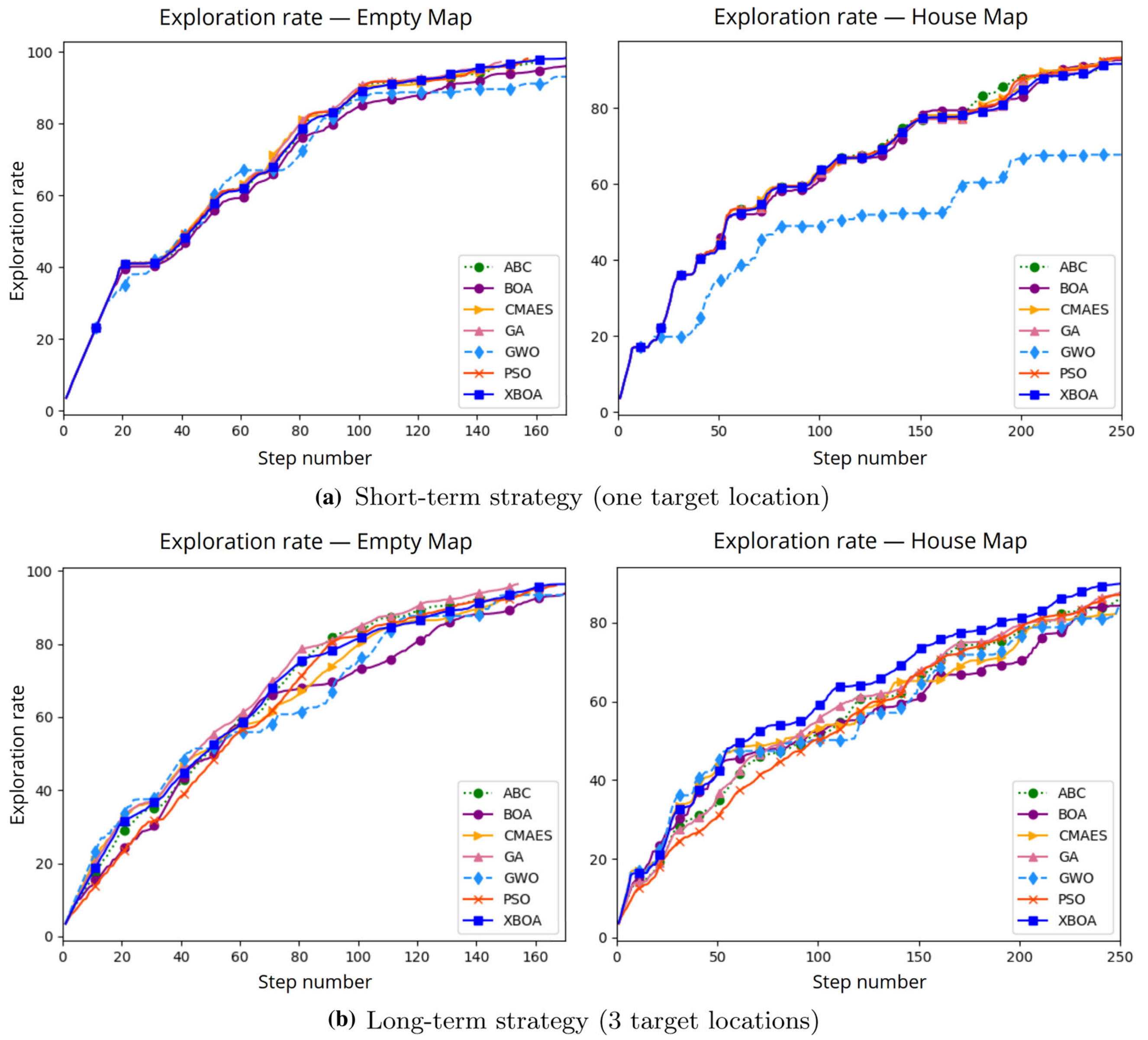

**Fig. 5** Simulation results of the exploration mission (average results for 10 runs)

Table 2 summarizes the best parameters found for each method after 90 trials using the Hyperopt library Bergstra et al. (2013). Some methods such as ABC, GWO, and CMAES require no manual tuning. The following parameters have been used for the other methods:

- GA: we used a Tournament selection strategy of size 2 and a polynomial mutation with a distribution index of 76.
- PSO: we used a neighborhood size of 4 with a cognitive acceleration factor bigger than the social acceleration factor, which means that the particles are more attracted toward the best position in their neighborhood than their previous best positions.
- The parameters of BOA-derived methods are explained in Sect. 3.

## 5.4 Results and discussion

### 5.4.1 Comparison between different optimizers

During the first batch of experiments, we compare the different metaheuristics using the short-term and long-term exploration strategies. As we can see from the results in Figure 5 and Table 3, both xBOA, PSO, GA, and CMAES perform similarly in most scenarios, especially when using the short-term strategy. This means that the optimization process successfully converged toward the optimal solution in most cases. However, we observe a decrease in the convergence of BOA and GWO in certain periods when the optimizer stuck in a local optimum, pushing the robot to revisit an already explored region for a certain amount of time. But it finishes by overcoming this local optimum and catching up with the other metaheuristics after a sufficient number of steps.

**Table 3** Exploration rates at the end of the mission

| Method | Empty map | | | House map | | |
|---|---|---|---|---|---|---|
| | Short-term exploration | | | | | |
| | Average | Min | Max | Average | Min | Max |
| *ABC* | 98.12 | 97.22 | ≥ **99** | 90.1 | 88.36 | 92.7 |
| BOA | 96 | 94.79 | ≥ **99** | **93.4** | 91.84 | 95.13 |
| CMAES | **98.49** | 95.66 | ≥ **99** | 93.29 | 89.4 | **96.18** |
| GA | 97.38 | 82.81 | ≥ **99** | 93 | **92.01** | 94.96 |
| GWO | 93.03 | 92.88 | 93.05 | 67.7 | 67.7 | 67.7 |
| PSO | 98.16 | **97.39** | ≥ **99** | 93.19 | 88.71 | 94.79 |
| xBOA | 98.23 | 93.92 | ≥ **99** | 91.63 | 82.63 | 94.44 |
| | Long-term exploration | | | | | |
| | Average | Min | *Max* | Average | Min | *Max* |
| ABC | **98.02** | **96.52** | ≥ **99** | 86.38 | 80.38 | 92.18 |
| BOA | 93.87 | 91.49 | 97.74 | 84.61 | 73.26 | 92.36 |
| CMAES | 94.16 | 83.85 | ≥ **99** | 82.23 | 77.43 | 86.45 |
| GA | 97.63 | **96.52** | ≥ **99** | 87.67 | 79.86 | **95.48** |
| GWO | 93.4 | 93.4 | 93.4 | 83.68 | 83.68 | 83.68 |
| PSO | 96.44 | 92.88 | ≥ **99** | 87.06 | **84.2** | 90.97 |
| xBOA | 96.37 | 93.57 | 98.61 | **89.9** | 79.16 | 94.1 |

It might be important to note that the “step number” here corresponds to the number of moves and rotations performed by the robot. This number is correlated with the mission duration but not necessarily in a linear fashion since the time needed to perform a rotation is different from the time needed for moving one meter forward.

The reason we choose to analyze the number of steps instead of exploration time is the importance of the energy factor in robotics applications. The number of moves that a robot can perform is limited by the battery capacity. If in our case, the robot had a battery capacity enough for performing only 100 steps, for example, it would explore 80% of the Empty Map area using BOA instead of 85% using xBOA, PSO, or CMAES. Given this evaluation criterion, BOA is less efficient than the other metaheuristics unless the robot has no battery limitations.

Also, using the long-term strategy seems less efficient. The reason is that the optimizer considers only the limited information it has when choosing multiple goal locations, but these locations might not be optimal, especially in the beginning phase when we have no prior information about the position of obstacles. The short-term strategy focuses rather on generating only one goal, then moving toward it while updating the map, and regenerating a new goal after reaching the previous one. The quality of the solutions is better in that case, especially for dynamic environments where the positions of obstacles change. However, the optimizer will repeat the evolution process more often (i.e., at each target location), which will slow down the mission if this process takes a long time to execute since the robot does not move during this operation.

The execution time recorded in Fig. 6 includes only the computational time required for the optimization process. Thus, the time for moving the robot is set to 0 since this operation involves moving the motors and requires very little CPU time. However, in real-world situations, performing these moves take a fair amount of time, especially when the speed of the robot is reduced nearby obstacles as a safety measure, which will increase the duration of the overall exploration mission.

Although xBOA, PSO, ABC, and GA give better results in terms of exploration rate compared to BOA, they require more time for executing the optimization process. A reason for this might be the simplicity and the reduced number of operations in the algorithm logic of BOA compared to the other methods. The results presented in Fig. 7 reinforce this supposition; we clearly see that the average computation time of BOA is lower than xBOA, PSO, and GA, even when the number of calls to the fitness function (i.e. fitness evaluations) is equal or bigger.

We also observe that ABC requires a considerable amount of time in all experiments, which is twice the time required for GA and PSO. This is caused by a large number of fitness evaluations, which impacts the total computation time of the algorithm making it not a suitable method to use for real-world scenarios when the robots need to keep moving.

CMAES is dominating the other methods from the execution time perspective, but it is sometimes dominated by BOA in terms of the total exploration rate at the end of the mission. xBOA dominates BOA in all scenarios, while GWO dominates BOA sometimes in the amount of computation time required but does not dominate it in the exploration rate criteria.

From Fig. 7, we note that the average computation time for finding the best target goal is relatively long, averaging 150∼450 seconds in which the robot is in an idle state waiting for the optimization result. This is not recommended in scenarios where the time factor is critical, such as in Research & Rescue missions. Two potential solutions are possible in this case: either to take advantage of parallelism and evaluate several candidate solutions simultaneously, or reduce the size of the population, which means reducing the number of candidate solutions to evaluate. In the next experiment, we analyze the effects of this later strategy on the quality of the generated solutions.

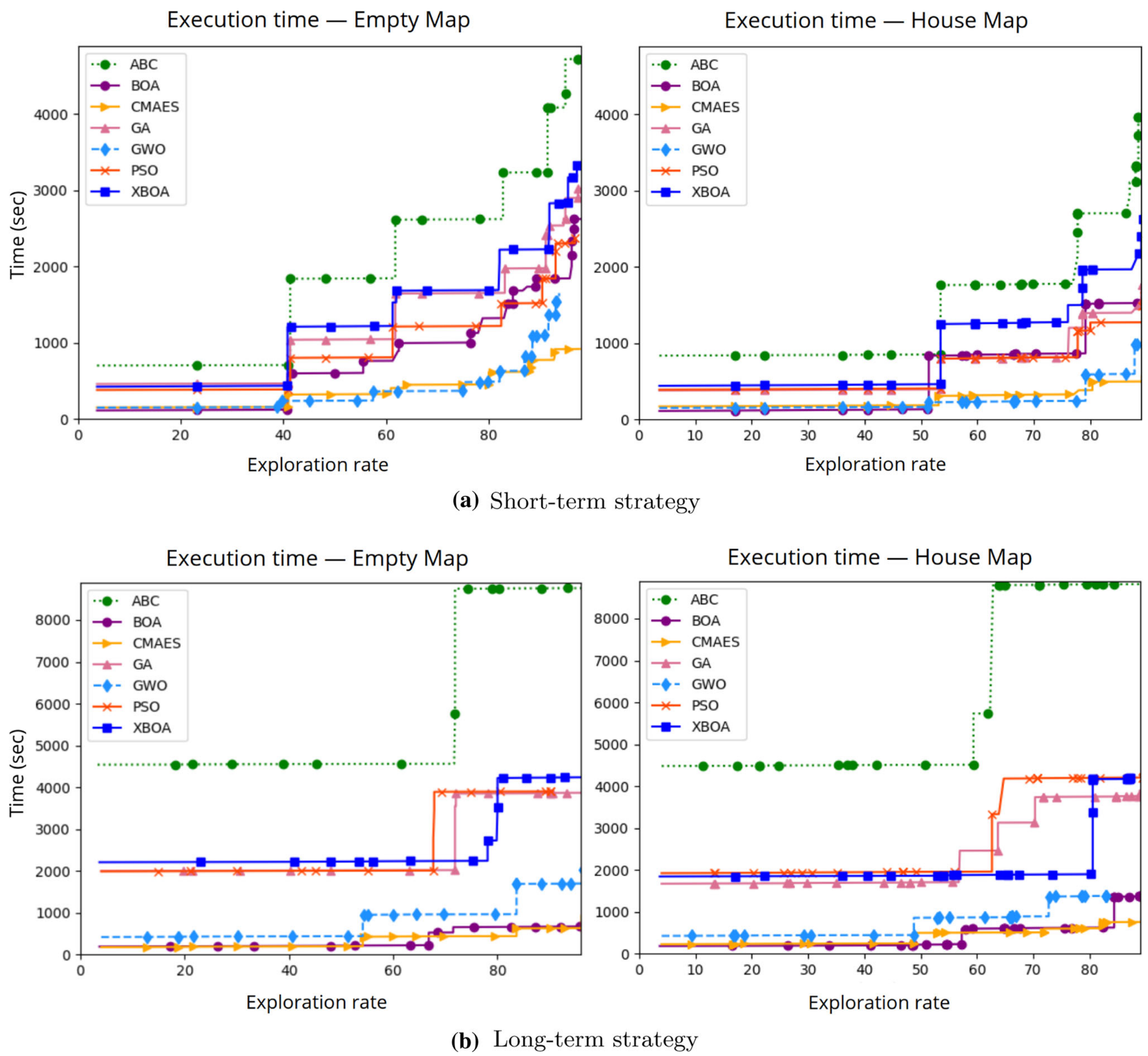

**Fig. 6** Comparison of the total execution time for the exploration mission (average results of 3 runs)

### 5.4.2 Evaluating the robustness to reduced population sizes

To study the impact of reducing the size of the population, we conducted a set of experiments varying this parameter from 20 to 5, while keeping the number of iterations unchanged. The results are reported in Figures 8 and 9.

We observe that using a population size of 5 individuals has a drastic improvement on the overall mission duration. However, it affects the quality of the exploration since the total explored area at the end of the mission using BOA on the House map dropped by 6.73% as we can see in Tables 4 and 5. The average duration for computing the next goal location is reduced to 110 s, which is acceptable for many robotics applications in the real world.

**Table 4** Exploration rates using BOA with different population sizes

| Short-term exploration - House map | | | |
|---|---|---|---|
| | Average | Min | Max |
| Pop size 05 | 86.11 | 81.77 | **95.13** |
| Pop size 10 | 88.66 | 79.68 | **95.13** |
| Pop size 20 | **92.84** | **89.75** | 94.27 |

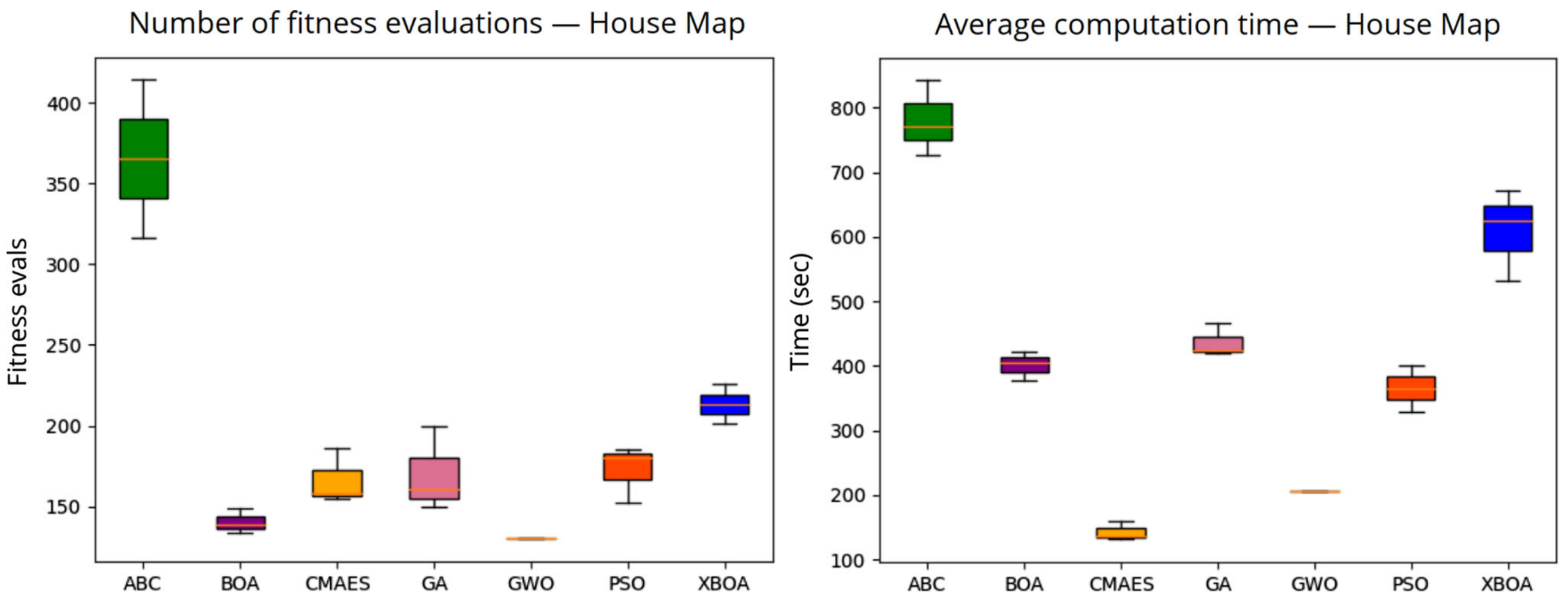

**Fig. 7** Number of fitness evaluations and average computation time (average results of 3 runs)

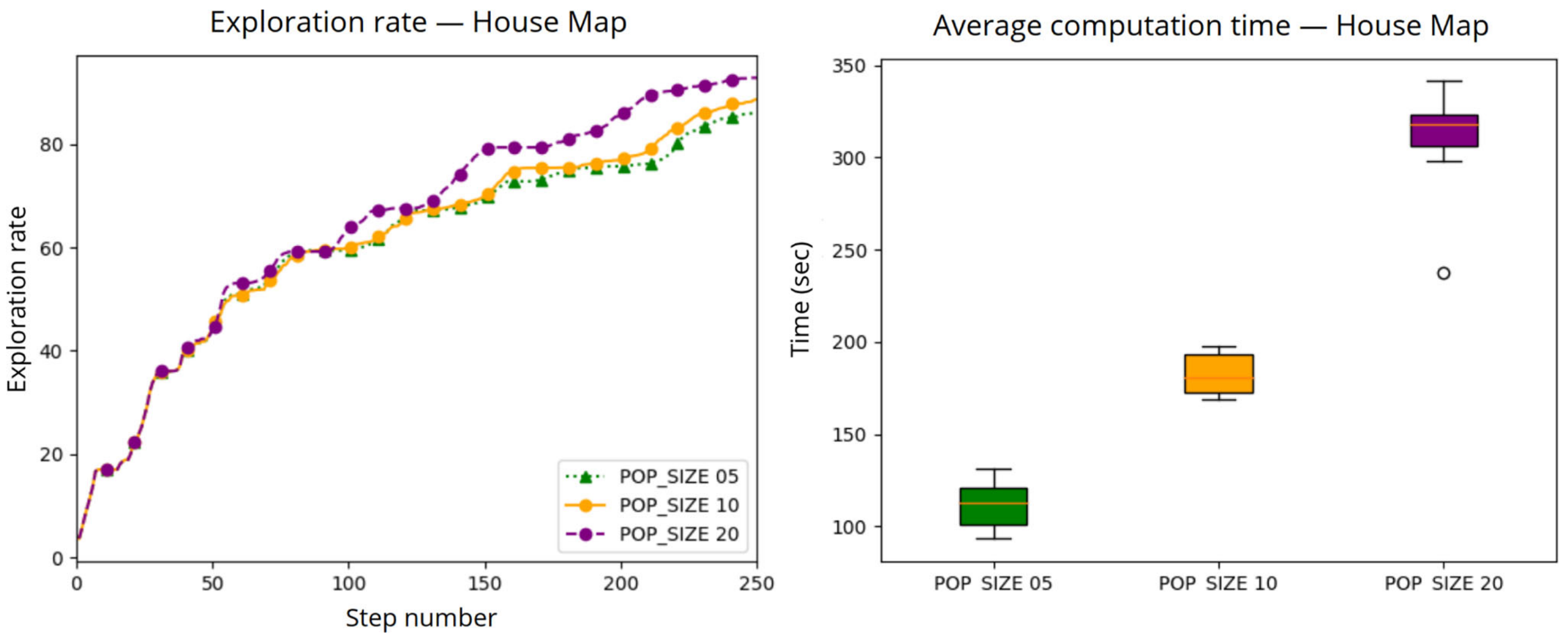

**Fig. 8** Comparison between different population sizes using BOA (average results of 3 runs)

We note that the PSO and GWO stopped improving after reaching 75 and 62% of the exploration rate, respectively, although they were showing results close to GA and ABC near the end of the mission when the population size was bigger. xBOA dominates all the methods in the exploration rate and the fitness value convergence, but CMAES still dominates them in the processing time; its average duration for computing the next goal location is reduced to less than 25 seconds, while ABC requires 175 seconds to get the same result.

All the optimizers stopped improving near the end of the exploration mission. The small number of individuals does not allow the algorithm to escape the local optima. One solution might be to increase the search space at the end of the exploration mission by gradually increasing the population size or increasing the number of iterations. That would be a better strategy than starting the whole mission with a large population size, which will consume a lot of time for almost the same results as shown in the previous experiments.

### 5.4.3 Comparing BOA variants

The next experiment compares the results of several variants of BOA. As we can see from Figure 10 and Table 6, introducing the crossover operator results in a better exploration rate and fitness convergence compared to the original BOA and its other variants. We can conclude then that the crossover operator helps in avoiding local minima. However, by creating new offspring the number of fitness evaluations increases, which slows the optimization process, even if the total duration of the exploration mission is not not significantly longer from the original BOA near the end of the mission.

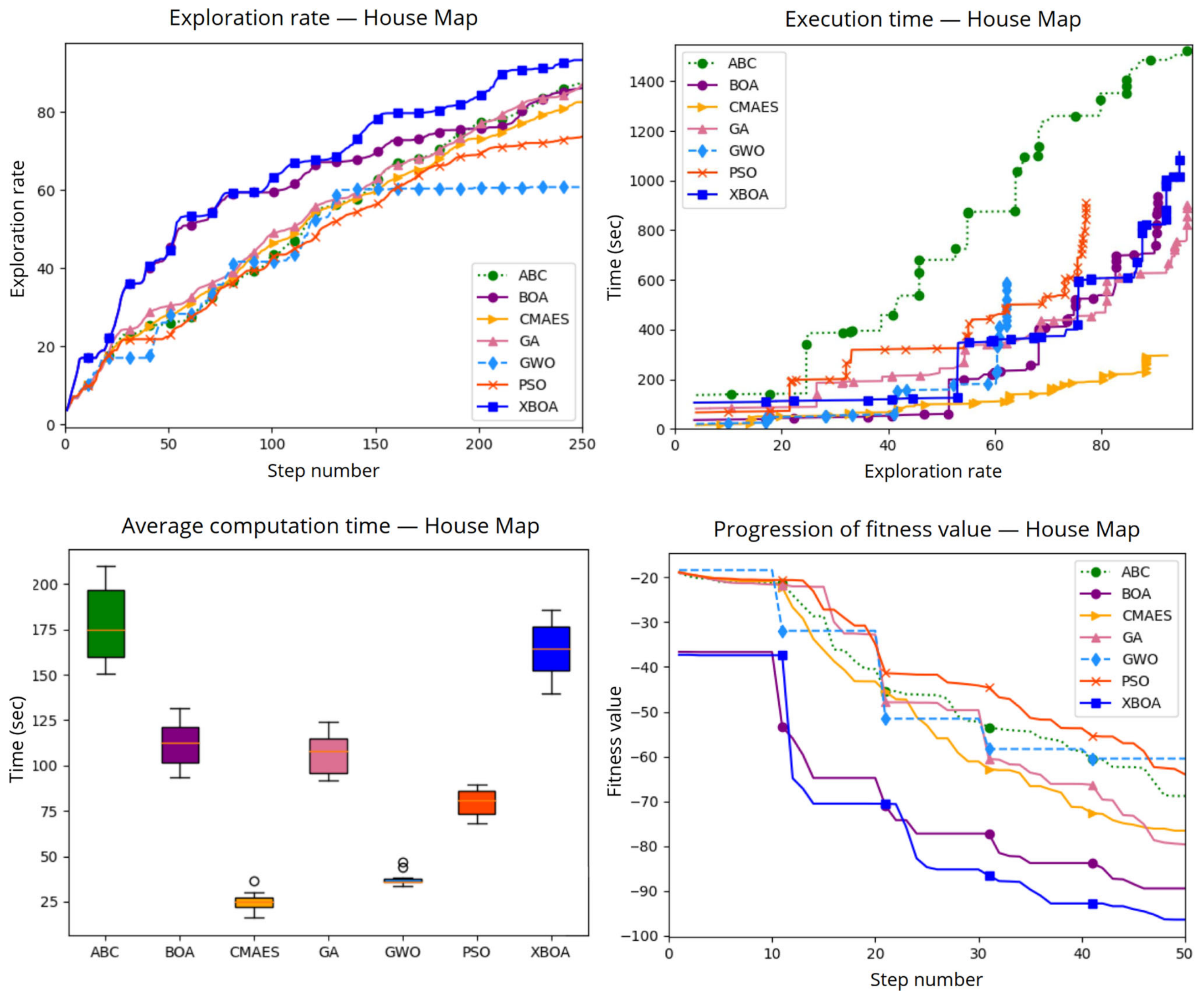

**Fig. 9** Exploration results using the short-term strategy and a population size of 5 individuals (average results of 3 runs)

While xBOA outperforms the other variants, an interesting alternative for emergency robotics applications would be ABOA that gives acceptable results in less time. It is ∼ 40% faster than xBOA and still gets better results than the original BOA. On the other hand, SABOA performs poorly, but it is easier to tune because of its auto-adaptative feature which reduces the number of hyperparameters to one. It is consequently more robust to bad initializations than the other variants.

### 5.4.4 Testing the multi-robots scenario

The final experiment intends to validate the adaptability of the approach for a multi-robot scenario. As we can see in Fig. 11, the robots successfully spread in the environment to explore the entire area although no changes have been made to explicitly coordinate their movements. This shows the flexibility of the proposed approach to adapt to a multi-robot scenario without any changes in the code since the proposed decision-making process is agnostic to the number of robots.

Each robot has its own population of candidate solutions and tries to maximize its own fitness reward regardless of the other robots. They do not exchange explicit messages between them and collaborate passively by modifying a shared occupancy map stored in a central memory. This is known as implicit coordination and has many advantages. However, it is not the optimal strategy because several robots may choose to go toward the same direction if it maximizes their fitness reward even if the other robots has already chosen it. To avoid this issue, we can modify the fitness function in a way to penalize overlapping paths by including information about current robots' positions and their target goals.

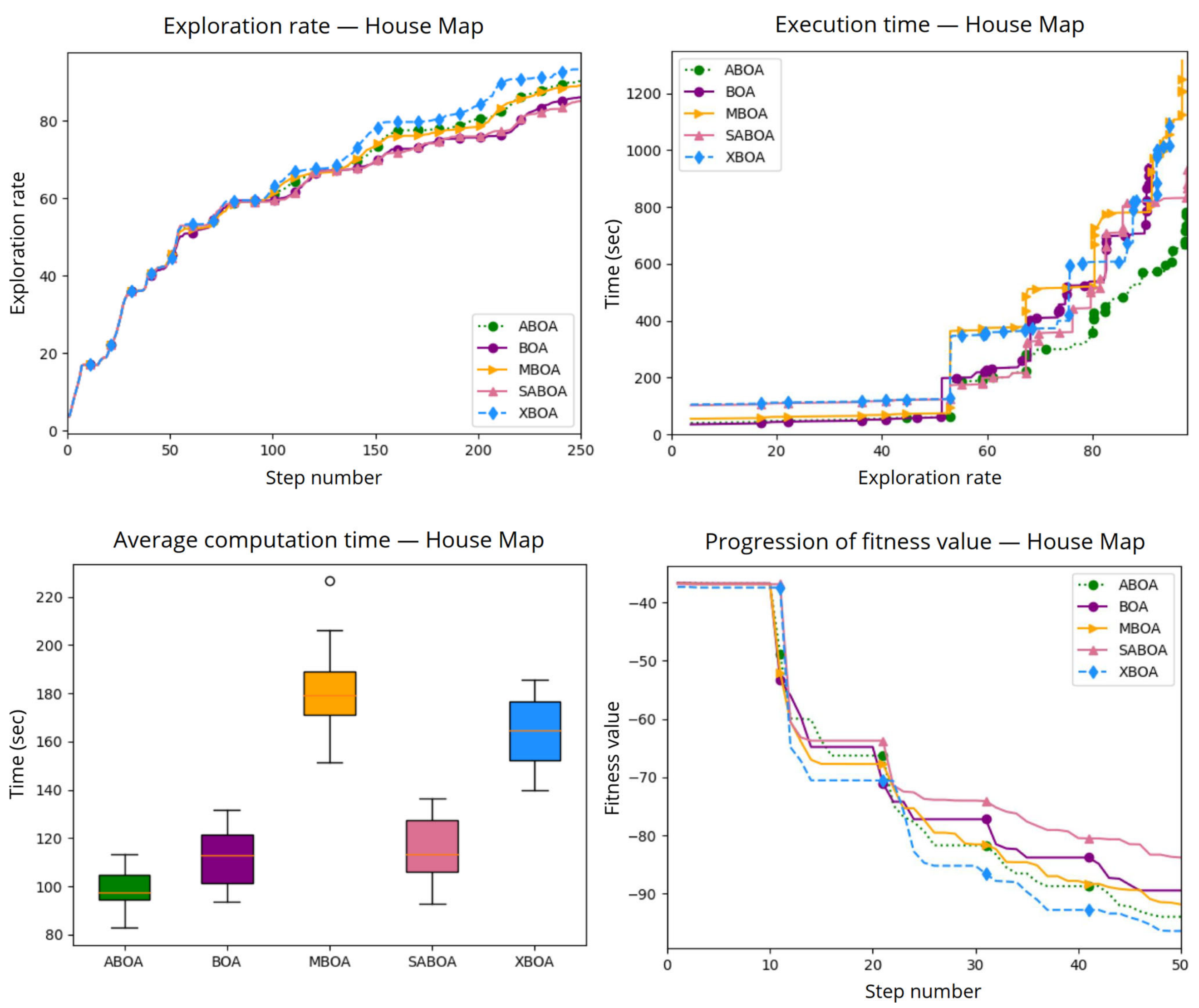

**Fig. 10** Comparing BOA variants using the short-term strategy and a population size of 5 individuals (average results of 10 runs)

We leave this solution for future investigations as it is outside the scope of our current work.

## 6 Conclusion

In this paper, we presented an approach for solving the Unknown Area Exploration problem using robots with limited energy. We proposed a new framework to solve this kind of problems incrementally using metaheuristics as optimizers for generating new target locations.

We used the Butterfly Optimization Algorithm (BOA) as a case study and adapted it to this problematic, and proposed a new variant called xBOA based on the crossover operator. We compared the two variants with several metaheuristics available in the literature through 5 different performance measures: steps number, mission duration, exploration rate, number of fitness evaluations, and the average computation time of the technique.

The results showed that xBOA outperforms the original BOA and its other variants, but requires a longer execution time. It also performs better than other metaheuristics such as PSO, GA, GWO, and ABC in some scenarios.

Finally, we tested our approach in a multi-robot setting without applying any changes to the model. The results demonstrated the adaptability of the proposed approach to both single- and multi-robot scenarios.

In future work, we will modify the fitness function to take advantage of multi-objective optimization strategies and extend the coordination capabilities of the robots by allowing them to exchange information about their positions and goal locations. We will also test the approach on real robots to evaluate the robustness of our simulation framework to real-world noisy environments.

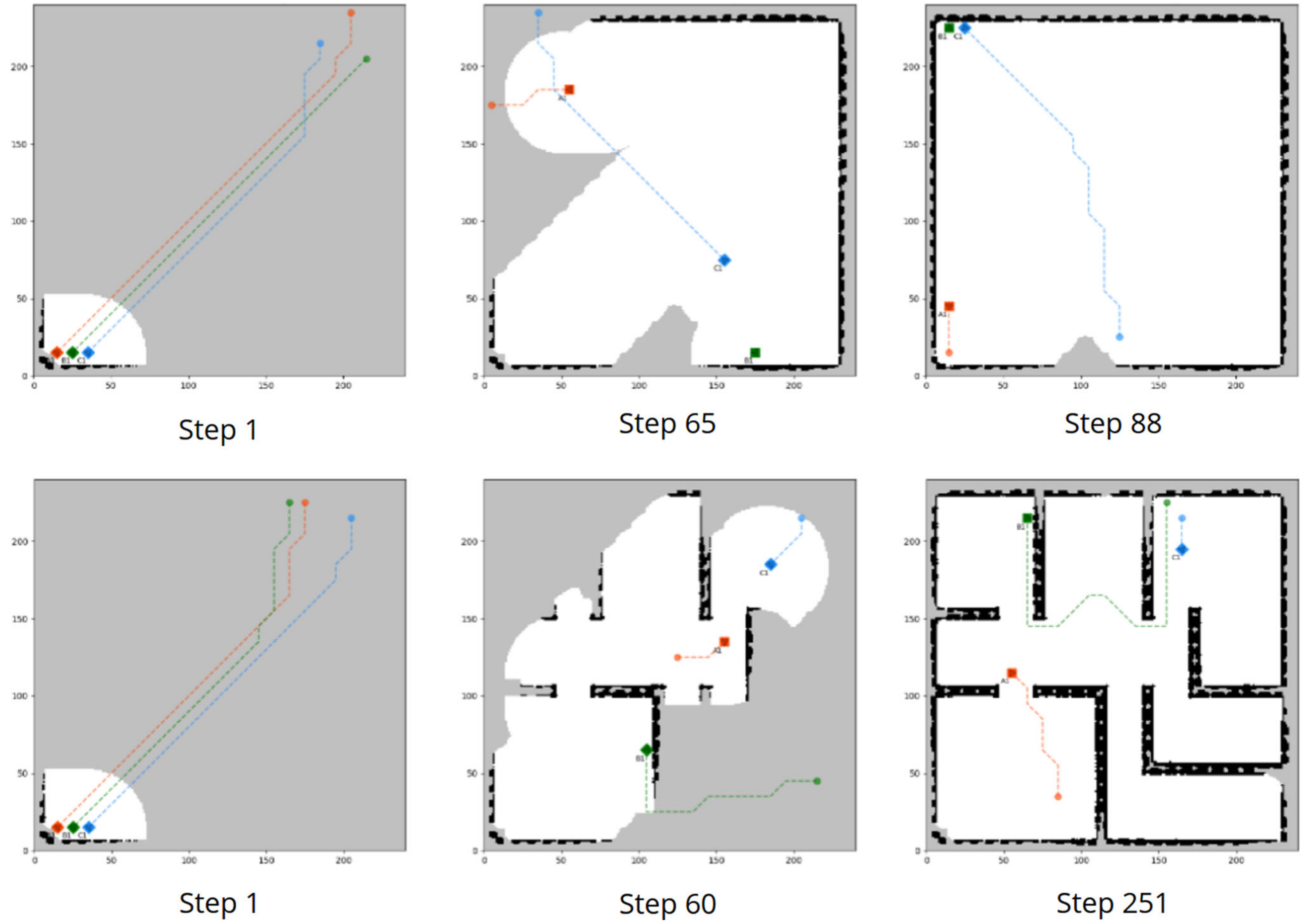

**Fig. 11** Visual progress of the multi-robot scenario using xBOA with 3 robots. Top: empty map experiment. Bottom: house map experiment

Another direction to improve this work is to investigate the limitations of BOA and xBOA. We might consider changing the switching strategy of the algorithm, testing different types of crossover operators, or applying xBOA to other types of global optimization problems. On another side, an extensive study has to be done to improve the theoretical aspects of the algorithm.

**Table 5** Exploration rates at the end of the mission using a population size of 5 individuals

| Short-term exploration | | | |
|---|---|---|---|
| | Empty map | | |
| Method | Average | Min | *Max* |
| ABC | **97.32** | **95.83** | 98.61 |
| BOA | 96.09 | 92.01 | $\geq$ **99** |
| CMAES | 94.46 | 89.4 | 98.61 |
| GA | 96.09 | 87.5 | $\geq$ **99** |
| GWO | 89.61 | 84.37 | 92.53 |
| PSO | 93.03 | 84.37 | 98.26 |
| xBOA | 96.14 | 92.88 | 98.26 |
| | House map | | |
| Method | Average | Min | *Max* |
| ABC | 90.59 | 84.72 | 96 |
| BOA | 87.13 | 83.68 | 95.13 |
| CMAES | 82.8 | 76.73 | 93.22 |
| GA | 92.36 | 89.23 | 96 |
| GWO | 62.03 | 60.76 | 62.32 |
| PSO | 74.9 | 70.83 | 77.6 |
| xBOA | **93.35** | **92.18** | **93.92** |

Bold indicates the maximum value of each column, which emphases the best result

**Table 6** Exploration rates using BOA variants with a population size of 5 individuals

| Short-term exploration - House map | | | |
|---|---|---|---|
| | Average | Min | Max |
| *BOA* | 89.11 | 81.77 | **95.13** |
| ABOA | 90.38 | 81.77 | **95.13** |
| mBOA | 89.14 | 81.07 | **95.13** |
| SABOA | 85.13 | 82.98 | 85.93 |
| xBOA | **93.35** | **92.18** | 93.92 |

Bold indicates the maximum value of each column, which emphases the best result

**Funding** No funds, grants, or other support was received.

**Data availability** Data and code are available at the following link: https://github.com/amineHorseman/butterfly-optimization-algorithms.

## Declarations

**Conflict of interest** The authors have no conflicts of interest to declare that are relevant to the content of this article.